\documentclass{article}
\usepackage{xcolor}         % colors
\definecolor{nipsblue}{rgb}{0.12, 0.49, 0.85}
\usepackage[colorlinks=true, linkcolor=nipsblue, citecolor=nipsblue, urlcolor=nipsblue]{hyperref}       % hyperlinks
\usepackage[preprint]{neurips_2023}
\usepackage{orcidlink}
\usepackage[utf8]{inputenc} % allow utf-8 input
\usepackage[T1]{fontenc}    % use 8-bit T1 fonts
\usepackage{url}            % simple URL typesetting
\usepackage{booktabs}       % professional-quality tables
\usepackage{amsfonts}       % blackboard math symbols
\usepackage{nicefrac}       % compact symbols for 1/2, etc.
\usepackage{microtype}      % microtypography
\usepackage{graphicx}
\usepackage{booktabs}
\usepackage{makecell}
\usepackage{multirow}
\usepackage{amsmath}
\title{Any2Point: Empowering Any-modality Large Models\\for Efficient 3D Understanding}

\author{
Yiwen Tang\textsuperscript{*\rm 1}, 
Ray Zhang\textsuperscript{*\rm 2},
Jiaming Liu\textsuperscript{*\rm 3},
Zoey Guo\textsuperscript{*\rm 2},
Dong Wang\textsuperscript{\rm 1}\\
\textbf{Zhigang Wang\textsuperscript{\rm 1},
Bin Zhao\textsuperscript{\rm 1,4},
Shanghang Zhang\textsuperscript{\rm 3},
Peng Gao\textsuperscript{\rm 1},
Hongsheng Li\textsuperscript{\rm 2},
Xuelong Li\textsuperscript{\rm 1}}\vspace{0.15cm}\\
\textsuperscript{\rm 1}Shanghai AI Lab
\textsuperscript{\rm 2}The Chinese University of Hong Kong\\
\textsuperscript{\rm 3}Peking University
\textsuperscript{\rm 4}Northwestern Polytechnical University \\
}
\begin{document}
\maketitle

\begin{abstract}
Large foundation models have recently emerged as a prominent focus of interest, attaining superior performance in widespread scenarios. Due to the scarcity of 3D data, many efforts have been made to adapt pre-trained transformers from vision to 3D domains. However, such 2D-to-3D approaches are still limited, due to the potential loss of spatial geometries and high computation cost. More importantly, their frameworks are mainly designed for 2D models, lacking a general any-to-3D paradigm. In this paper, we introduce \textbf{Any2Point}, a parameter-efficient method to empower any-modality large models (vision, language, audio) for 3D understanding. Given a frozen transformer from any source modality, we propose a 3D-to-any (1D or 2D) virtual projection strategy that correlates the input 3D points to the original 1D or 2D positions within the source modality. This mechanism enables us to assign each 3D token with a positional encoding paired with the pre-trained model, which avoids 3D geometry loss caused by the true projection and better motivates the transformer for 3D learning with 1D/2D positional priors. Then, within each transformer block, we insert an any-to-3D guided adapter module for parameter-efficient fine-tuning. The adapter incorporates prior spatial knowledge from the source modality to guide the local feature aggregation of 3D tokens, compelling the semantic adaption of any-modality transformers. We conduct extensive experiments to showcase the effectiveness and efficiency of our method. The code is released at \url{https://github.com/Ivan-Tang-3D/Any2Point}.
\end{abstract}

\section{Introduction}
\label{sec:intro}
Driven by the growing volume of model parameters and training data, large foundation models have gained unprecedented attention in a diverse array of domains and tasks. Numerous large models have been pre-trained for natural language process, including BERT \cite{devlin2018bert}, T5 \cite{raffel2020exploring}, and GPT series \cite{openai2023gpt4, floridi2020gpt}, as well as visual understanding like DINOV2 \cite{oquab2023dinov2}, MAE \cite{he2022masked, wei2022masked, xie2022simmim}, and ViT-22B \cite{dehghani2023scaling}. Existing works \cite{hu2021lora, liu2021p, chen2022adaptformer, jia2022visual} also explore efficient fine-tuning techniques to transfer pre-trained large models to a variety of downstream tasks, consistently achieving excellent performance. Meanwhile, 3D visual understanding \cite{zhang2022pointclip, dai2017scannet, qi2017pointnet, guo2023point} is also a significant topic, with its rich geometric representation contributing to the development of many applications (e.g., robotics \cite{li2023manipllm,guo2023viewrefer} and autonomous driving \cite{yang2023lidar, pan2023renderocc}). Unfortunately, due to a lack of large-scale 3D data, the efforts towards 3D foundational modal are significantly lagging compared to language and 2D vision. Specifically, the acquisition and annotation of high-quality 3D data requires expensive resources and human labor, while synthetic 3D data training falls short of distribution diversity and real-world applications.

\begin{figure*}[t!]
\vspace{-0.1cm}
\centering
\includegraphics[width=0.95\textwidth]{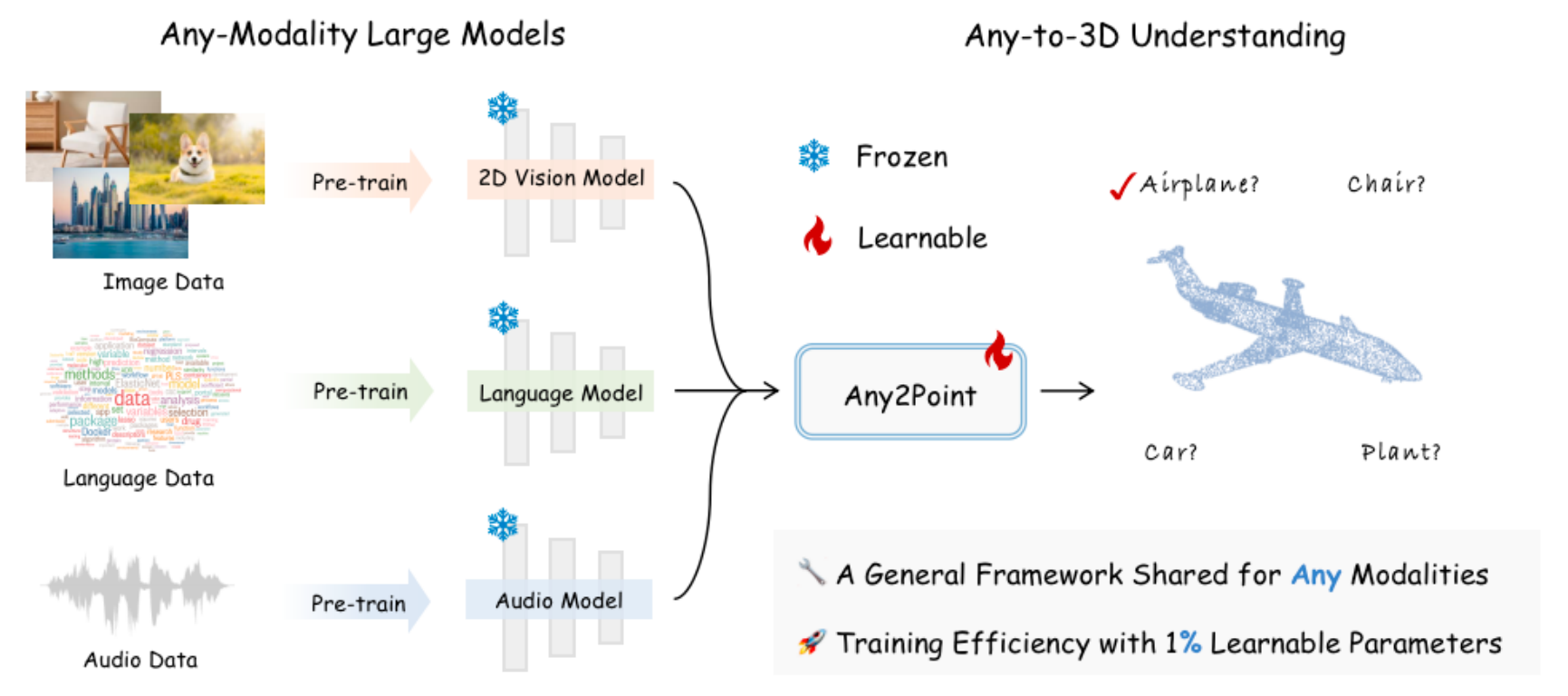}
   \caption{\textbf{Overview of Any2Point.} We propose a general framework for any-to-3D learning, which is shared for any modalities with parameter-efficient fine-tuning.}
\label{intro}
\end{figure*}

Therefore, some previous works have transferred pre-trained models from other modalities (mainly 2D vision) to 3D modality, leveraging sufficient pre-trained knowledge from diverse sources. We categorize existing 2D-to-3D works into two groups. \textbf{1) Data modality transformation.} This type of approach involves projecting 3D point clouds into 2D images \cite{wang2022p2p, zhang2022pointclip, zhu2023pointclip}, which are subsequently fed into 2D pre-trained models. Despite the promising performance on downstream tasks, the process of modality transformation inevitably causes the loss of spatial information in 3D data, hindering the full potential for 3D understanding. \textbf{2) Cross-modality knowledge distillation.} These approaches involve the pre-training knowledge transfer from 2D or vision-language models to a newly trained 3D model \cite{zhang2023learning, dong2022autoencoders, xue2023ulip}. They are not only required to forward propagate both the 2D and 3D models during training, but also highly rely on the large-scale paired 2D-3D data. This leads to substantial computation costs and data engineering, limiting their ability for efficient implementation. Besides the aforementioned issues, more importantly, current methods mostly focus on the model adaption from 2D vision to 3D point clouds, rather than a shared methodology for other modalities.
Therefore, we pose a question: \textit{can we develop a general any-to-3D paradigm that empowers any-modality large models for efficient and effective point cloud understanding?}

To address this issue, we propose Any2Point, a unified any-to-3D framework that transfers any 1D (language) or 2D (image/audio) large models to 3D domains with Parameter-Efficient Fine-Tuning (PEFT), as shown in Figure~\ref{intro}. Different from prior methods, our Any2Point avoids the point cloud projection, thereby mitigating the 3D information loss, and directly fine-tunes pre-trained models from source modalities, which saves resources by knowledge distillation. Specifically, given an any-modality pre-trained transformer, we first introduce a 3D-to-any (1D or 2D) virtual projection mechanism. This mechanism establishes a positional mapping between the input 3D points and their virtually projected 1D lines or 2D planes. This enables us to encode 3D coordinates using the original positional embeddings of the source modality of pre-trained large models.
In this way, we no longer need to conduct a true projection losing 3D geometries, while better promoting the pre-trained transformer to acquire 3D features with their original 1D/2D positional priors.
Then, for each transformer block, we insert an any-to-3D guided adapter module for PEFT. This adapter leverages the 1D/2D spatial guidance to aggregate the local semantics of 3D tokens, facilitating fine-grained feature interaction. Afterward, we perform an adaptive ensemble for the 3D features guided by different 1D/2D priors, which attains superior 3D representations.

Extensive experiments across various tasks demonstrate that our Any2Point framework achieves superior performance compared to current 3D pre-trained models, while utilizing only 1.0\% of the trainable parameters. Using the pre-trained CLIP Text Encoder \cite{radford2021learning}, Any2Point fine-tunes only 0.8M parameters and attains 91.9\% on ScanObjectNN \cite{uy2019revisiting}, outperforming the previous state-of-the-art (SOTA) 3D pre-trained model by +1.3\%, and 94.3\% on ModelNet40 \cite{wu20153d}. Furthermore, Any2Point also achieves comparable results and efficiency by utilizing other pre-trained models \cite{oquab2023dinov2,touvron2021training,gong2022ssast,girdhar2023imagebind,liu2019roberta} of different modalities, including 2D vision, language, and audio, validating the robustness of our approach. The contributions of our paper are as follows:
\begin{itemize}
\item 
% We propose a unified cross-modality PEFT framework, named Any2Point, transfering pre-trained large models from any modalities (i.e., language, 2D image and audio) to 3D point cloud tasks.
To enable a general any-to-3D transferring framework, we propose Any2Point, which empowers any-modality pre-trained large models (e.g., 2D vision, language, and audio) for efficient 3D understanding.
\item
% We design a Virtual Coordinate Projection (VCP) mechanism to encode 3D point cloud data directly using pre-trained positional embeddings from large-scale models. Additionally, we introduce a Spatial-aware Adapter (SaA) to efficiently extract both global and local 3D spatial knowledge.
We introduce two techniques, i.e., 3D-to-any virtual projection and any-to-3D guided adapter, to effectively overcome the issues within current methods, such as 3D geometry loss and excessive resource cost.

% \item
% We propose an any-to-3D guided adapter to compel the semantic adaption of any-modality transformers by utilizing 1D/2D positional priors. 

\item
Any2Point achieves superior performance compared to previous SOTA 3D pre-trained models across various tasks. Notably, these competitive results remain consistent by leveraging pre-trained models from different modalities, e.g., 2D vision, language, and audio.

\end{itemize}

\section{Related Work}
\label{rw}
\subsection{Large Models}
Large-scale pre-trained models have achieved remarkable downstream performance in language, 2D vision, and audio processing. In the field of natural language field, BERT~\cite{devlin2018bert} revolutionized natural language understanding by pre-training deep bidirectional representations from unlabeled text. 
Building on this, RoBERTa~\cite{liu2019roberta} enhances BERT by optimizing training and data scale, significantly boosting performance across language understanding benchmarks. GPT~\cite{floridi2020gpt,radford2019language} pioneers in generating coherent, contextually relevant text using a transformer model pre-trained on a diverse corpus of text. CLIP~\cite{radford2021learning} further bridges visual and linguistic information by pre-training on a vast collection of image-text pairs.
In 2D vision, DeiT~\cite{touvron2021training} achieves efficient image classification with transformers, using data augmentation and knowledge distillation for minimal data reliance. DINO V2~\cite{oquab2023dinov2} also advances self-supervised learning with vision transformers through innovative self-distillation, requiring no labels. MAE~\cite{he2022masked} further proposes an asymmetric encoder and decoder to reconstruct images from masked data.
For audio processing, AST~\cite{gong2021ast} transforms audio recognition to spectrograms by applying a vision transformer. SSAST~\cite{gong2022ssast} leverages self-supervised learning on unlabeled data with transformers for enhanced audio classification performance. ImageBind~\cite{girdhar2023imagebind} unifies multi-modal space by jointly training on multi-modal data including audio, improving multi-modal understanding and generation. Our method first utilizes the abundant knowledge from large models of any modalities and achieves 3D understanding capacity.

\subsection{2D-to-3D Transfer Learning}
The paradigm of 2D-to-3D transfer learning aims to leverage the rich contextual and textural knowledge in the 2D domain to boost 3D understanding. Some works propose specific designs for 3D learning guided by 2D pre-trained knowledge, and achieve promising 3D understanding performance.
Image2Point~\cite{xu2022image2point} proposes to transfer 2D semantics to 3D by convolutional layer inflating. ACT~\cite{dong2022autoencoders} employs pre-trained 2D and language Transformers as cross-modal teachers for 3D learning via discrete variational autoencoding and prompt tuning. ULIP~\cite{xue2023ulip} enhances 3D understanding performance by unifying image, text, and 3D point cloud representations. ReCon~\cite{qi2023contrast} leverages contrastive cross-modal learning and generative models for knowledge transfer. Meanwhile, PointCLIP V1~\cite{zhang2022pointclip} and V2~\cite{zhu2023pointclip} first adopt CLIP’s 2D pre-trained knowledge on different 3D downstream tasks via projecting 3D point clouds to 2D images as input to the pre-trained backbone. P2P~\cite{wang2022p2p} also proposes 2D-to-3D projection through a learnable coloring module. Our approach skips the step of projecting point clouds, which reduces the loss of 3D geometric information, and fine-tunes pre-trained models instead of using computationally expensive knowledge distillation.

\subsection{Parameter-Efficient Fine-tuning}
The pre-training and fine-tuning paradigm has been proven highly effective in various tasks such as 2D visual recognition, language understanding, text-to-image generation, and audio recognition. However, fully fine-tuning the whole model becomes impractical as model volume increases exponentially. In contrast, the parameter-efficient fine-tuning (PEFT) methods~\cite{hu2021lora,jia2022visual, chen2022adaptformer,houlsby2019parameter, liu2023vida,duan2024causal} aim at only updating a tiny part of the model's parameters while freezing the rest parts, which have been proven to be effective and efficient on multiple popular pre-trained models such as BERT~\cite{devlin2018bert},
GPT~\cite{floridi2020gpt}, ViT~\cite{dosovitskiy2020image}, CLIP~\cite{radford2021learning}, and Stable Diffusion~\cite{rombach2022high}. 
The PEFT approaches can be generally categorized into three principal streams
namely prompt tuning~\cite{jia2022visual, yang2024exploring}, reparameterization~\cite{hu2021lora}, and adapters~\cite{chen2022adaptformer,houlsby2019parameter,zhang2023llama}. These techniques customize pre-trained models for specific tasks by fine-tuning prompts, modifying model parameters without altering the original architecture, and inserting lightweight trainable layers, respectively. Recently, Point-PEFT~\cite{tang2024point} first introduced PEFT techniques into 3D domains. In this paper, the Any2Point framework utilizes PEFT techniques to transfer Any-Modality pre-trained models to 3D understanding tasks at a low computational and storage cost.

\section{Any2Point}

In Section~\ref{sec:Overview}, we first provide a paradigm overview of Any2Point, including the problem definition and network architecture. Then, in Section~\ref{sec:Virtual Projection} and~\ref{sec:Guided Adapter}, we respectively elaborate on the methodologies of our proposed two techniques for adapting any-modality large models for 3D domains.

\begin{figure*}[t!]
\vspace{-0.1cm}
\centering
\includegraphics[width=0.95\textwidth]{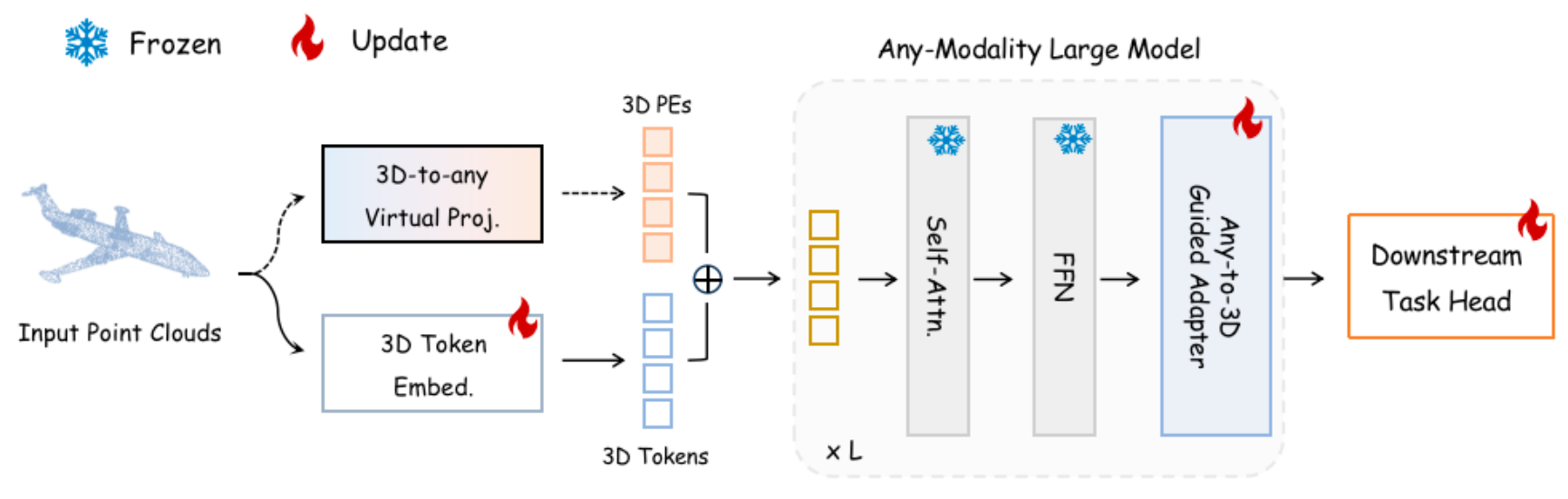}
   \caption{\textbf{Overall Pipeline of Any2Point.} For efficiently fine-tuning Any-modality pre-trained models, our Any2Point framework contains two components: a 3D-to-any Virtual Projection, which pairs the pre-trained positional encodings with 3D tokens to avoid the 3D geometric information loss, and an Any-to-3D Guided Adapter to effectively grasp local structures. }
\label{pipeline}
\vspace{-0.2cm}
\end{figure*}

\subsection{Method Overview}
\label{sec:Overview}
\paragraph{\textbf{Problem Definition.}} 
Given a pre-trained transformer from any modality, e.g., vision, language, and audio, our objective is to empower it with 3D understanding capabilities in an effective and efficient manner. Instead of employing full fine-tuning on 3D data, we seek a parameter-efficient solution with the source transformers frozen, since their large-scale parameters might cause high computation cost and over-fitting issues on the limited 3D dataset.
We generally divide the source models into two categories according to their pre-training data dimension, denoted as 1D and 2D transformers. The 1D transformers are specialized in processing sequential data, exemplified by language models like RoBERTa~\cite{liu2019roberta}, T5~\cite{raffel2020exploring}, and CLIP's text encoder~\cite{radford2021learning}. The 2D transformers are expert at 2D spatial data, including vision models, e.g., DINOv2~\cite{oquab2023dinov2} and DeiT~\cite{touvron2021training}, and audio models, e.g., ImageBind Audio Encoder~\cite{girdhar2023imagebind} and SSAST~\cite{gong2022ssast}.

\paragraph{\textbf{Model Pipeline.}}
The overall paradigm of Any2Point is depicted in Figure~\ref{pipeline}. To encode the input point cloud, we discard the original embedding modules in source transformers, e.g., tokenizers in 1D language models and convolutions in 2D vision/audio models, and employ a 3D mini-network for point cloud tokenization. 
On top of this, the encoded 3D tokens are fed first into a 3D-to-any virtual projection module for positional encoding, and then into the frozen 1D/2D transformer with any-to-3D guided adapters. The former mechanism aims to assign each 3D token with positional information within the source modality, and the latter is designed for adaptive 1D/2D-guided 3D representation learning, which we will detail in the following sections. Note that, as the source transformers are kept frozen, only the initial tokenization network and the inserted adapters are learnable for parameter-efficient fine-tuning.

% Specifically, the 3D mini-network is a lighter-weight variant modified from Point-PN~\cite{} to transform the raw point clouds into high-dimensional vectors. 
\begin{figure*}[t!]
\vspace{-0.1cm}
\centering
\includegraphics[width=0.95\textwidth]{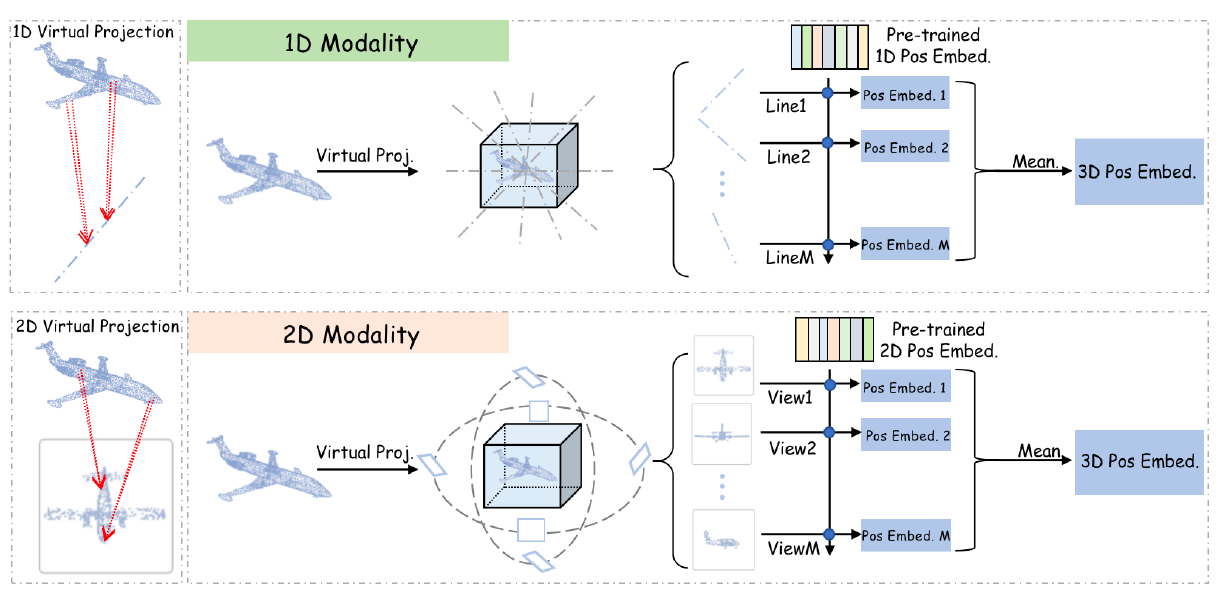}
   \caption{\textbf{3D-to-any Virtual Projection.} To prevent the loss of 3D geometric information, the module assigns 3D tokens with the positional encodings that are paired with the pre-trained model.}
\label{project}
\end{figure*}

\subsection{3D-to-any Virtual Projection}
\label{sec:Virtual Projection}
Many current 2D-to-3D methods~\cite{zhu2023pointclip,zhang2022pointclip,wang2022p2p} project 3D point clouds into multi-view images to meet the input modality of pre-trained 2D models. This dimension reduction process potentially leads to the information loss of 3D geometries and deep measurements, enabling insufficient 3D feature encoding. In addition, these approaches are merely validated on the large models within 2D images, without considering other modalities like language and audio. Therefore, we propose a 3D-to-any virtual projection strategy that mitigates the geometric loss, and is generalizable to any 1D/2D pre-trained models, as shown in Figure~\ref{project}.

\paragraph{\textbf{Tokenization in 3D Space.}}
To avoid any information degradation, we directly tokenize the input point cloud within the 3D space for the subsequent 1D/2D transformer. Specifically, we employ a 3D mini-network containing small-scale parameters, which is a lighter-weight variant of Point-PN~\cite{zhang2023starting,zhu2024no} . The tokenization process involves Farthest Point Sampling (FPS)~\cite{qi2017pointnet} for point number downsampling, $k$-Nearest Neighbor ($k$-NN) algorithm for local aggregation, and learnable linear layers for feature encoding. After this, we transform the raw point clouds into high-dimensional vectors, obtaining $N$ 3D tokens as $\{T_i\}_{i=1}^N$, with $\{p^{3D}_i\}_{i=1}^N$ denoting their 3D coordinates.

\paragraph{\textbf{Motivations for Virtual Projection.}}
Positional encodings (PEs) serve as the only indicator for positional information to the transformer model, since the inner attention mechanism is permutation-invariant, treating every token at different orders all the same. Therefore, a straightforward way for 1D/2D transformers to comprehend 3D positional information is to integrate new 3D PEs with 3D tokens. However, the source transformers are pre-trained paired with their original PEs in 1D/2D space, which leads to semantic discrepancy between the frozen 1D/2D weights and newly learned 3D PEs. To address this issue, we virtually project 3D tokens into the source modality, and obtain the corresponding 1D/2D PEs for better aligning with the transformers. 

\paragraph{\textbf{3D-to-2D Virtual Projection.}}

For 2D transformers in 2D vision and audio modalities, we virtually project each 3D coordinate, e.g., $p^{3D}_i$, into $M$ views, deriving the corresponding 2D coordinates as $\{p^{2D}_{ij}\}_{j=1}^M$. The $M$ different perspectives are capable of providing diverse positional relations within 2D space. We adopt a simple projection in PointCLIP~\cite{zhang2022pointclip} without learnable parameters. Importantly, we do not truly produce the projected multi-view images, but only aim to obtain the virtual 2D positions. Then, according to the original 2D PEs within pre-trained transformers, we assign each 3D token, e.g., $T_i$, with $M$ different PEs, denoted as $\{\operatorname{PE}^{2D}(p^{2D}_{ij})\}_{j=1}^M$.

\paragraph{\textbf{3D-to-1D Virtual Projection.}}
Similarly, for 1D transformers in language modality, we virtually project the 3D coordinates into different 1D lines. To align the number with 2D modality, we also select $M$ lines passing through the center of the point cloud with $M$ uniform rotation angles. For simplicity, we suppose the point cloud center as the origin, the unit direction vectors of $M$ lines as $\{\vec{v}^{1D}_{j}\}_{j=1}^M$, and the point coordinate, $p^{3D}_i$, vectorized as $\vec{p}^{3D}_i$. Then, the 1D coordinate of point $i$ in line $j$ is formulated by the dot production of
\begin{equation}
     p^{1D}_{ij} = \vec{v}^{1D}_{j}\cdot \vec{p}^{3D}_i,
\end{equation}
denoting the projected length. In this way, we refer to the original 1D PEs, and assign each 3D token, e.g., $T_i$, with $M$ different PEs as $\{\operatorname{PE}^{1D}(p^{1D}_{ij})\}_{j=1}^M$.

\paragraph{\textbf{Encoding 3D Positions in 1D/2D PEs.}}
After acquiring the corresponding 1D/2D PEs, we average them as an overall positional indicator, and incorporate it with the 3D token, e.g., $T_i$, by
\begin{equation}
     T_{i}^{in} = T_{i} + \frac{1}{M}\sum_{j=1}^M \operatorname{PE}^{1D/2D}(p^{1D/2D}_{ij}).
\end{equation}
With this approach, we inject sufficient positional information of the source modality into 3D tokens to better collaborate with the frozen transformer, while mitigating the information loss of the true projection.

\begin{figure*}[t!]
\centering
\includegraphics[width=0.94\textwidth]{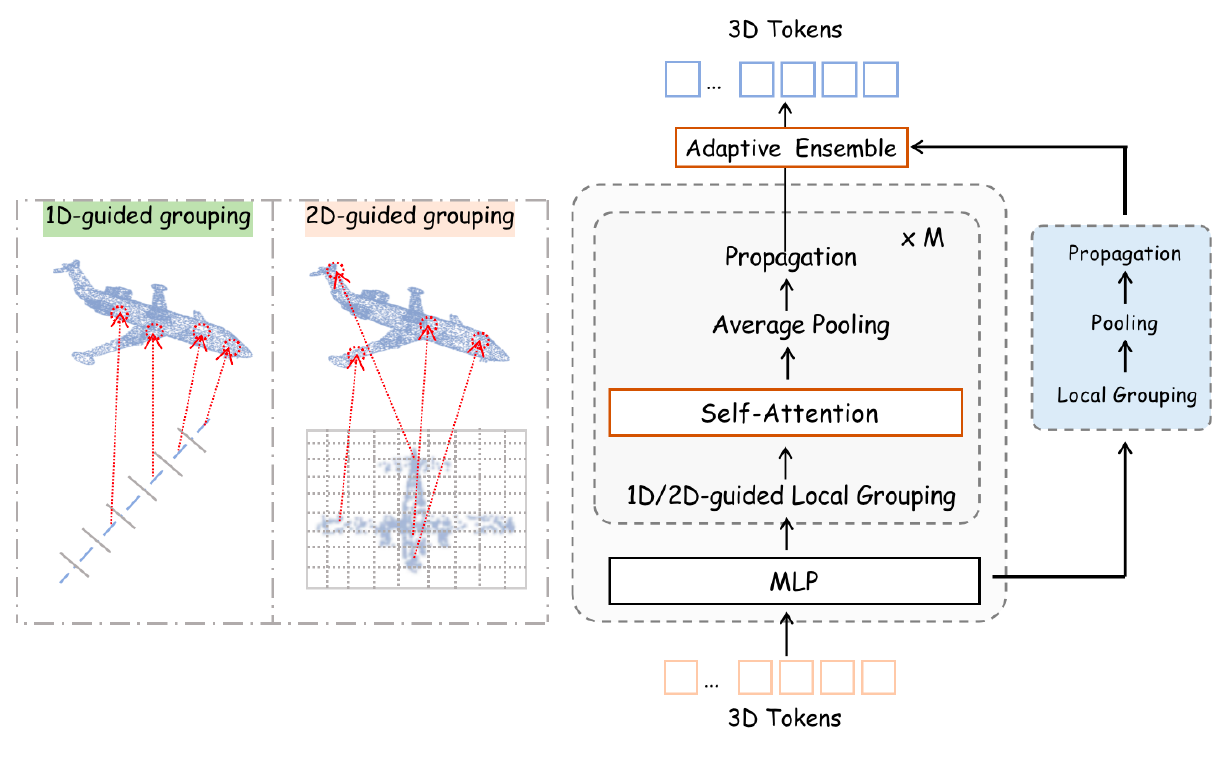}
   \caption{\textbf{Any-to-3D Guided Adapter.} Inserted into every transformer block, the adapter leverages the 1D/2D-guided Local Aggregation module to capture 3D local semantics and utilizes the Adaptive Any-to-3D Ensemble to obtain high-quality features.}
\label{adapter}
\vspace{-0.2cm}
\end{figure*}

\subsection{Any-to-3D Guided Adapter}
\label{sec:Guided Adapter}
Different from existing distillation-based methods~\cite{zhang2023learning,guo2023joint} training a new 3D network, we directly feed the encoded 3D tokens $\{T^{in}_{ij}\}_{i=1}^N$ to the pre-trained 1D/2D transformer. Although the PEs of 3D tokens have been aligned with the source model, the entirely frozen weights pre-trained by other modalities are still restricted to learning superior 3D representations. Considering this, we introduce a learnable any-to-3D guided adapter within each transformer block, as shown in Figure~\ref{adapter}. The adapters are inserted after the Feed-Forward Networks (FFNs), and further incorporate 1D/2D-prior knowledge for parameter-efficient fine-tuning.

\paragraph{\textbf{Motivations for Inserting Adapters.}}
The self-attention mechanisms within source transformers normally focus on long-range token interaction in global contexts, which lacks local feature extraction. However, the detailed spatial geometries are also significant for the fine-grained understanding of 3D shapes. To complement the gap, we utilize the proposed adapter layers for specifically capturing 3D semantics within local neighborhoods. In addition, as the source transformers are powered by 1D/2D PEs as discussed above, the naive FPS and $k$-NN for 3D local grouping might cause positional discrepancy. Therefore, we further design a 1D/2D-guided aggregation strategy and an adaptive any-to-3D ensemble approach for robust 3D fine-grained encoding.

\paragraph{\textbf{1D/2D-guided Local Aggregation.}}
Within the adapter, we first group 3D tokens into different local neighborhoods guided by 1D/2D positional priors, which better align the adopted 1D/2D PEs. For $M$ different views/lines, we conduct $M$ concurrent local aggregation process to make the best of different projection perspectives. Specifically, for 2D transformers, we divide each virtually projected image, e.g., the $j$-th view, into uniform local 2D patches, and group the 3D tokens within the same patch into a neighborhood, according to their 2D positions $\{p^{2D}_{ij}\}_{i=1}^N$. For 1D transformers, we similarly divide each virtually projected line, e.g., the $j$-th direction, into uniform local 1D segments, and group the 3D tokens within different segments referring to their 1D positions $\{p^{1D}_{ij}\}_{i=1}^N$. On top of this, we adopt a self-attention layer for 3D tokens within each 1D/2D neighborhoods, performing local feature interaction guided by 1D/2D priors. Then we employ the operations of pooling and propagation to propagate the local aggregated feature to every points within the same neighborhood.
% Note that, learnable weights of the attention layer are shared across $M$ views/lines for parameter efficiency.

\paragraph{\textbf{Adaptive Any-to-3D Ensemble.}}
After the parallel local aggregation, we obtain $M$ sets of 3D tokens, each representing a 2D view or 1D line. As different projection perspectives normally showcase different significance for 3D representations, we propose an adaptive any-to-3D ensemble approach to aggregate the $M$ features for each token. We denote the $i$-th 3D token with $M$ sets of features at this stage as $\{F_{ij}\}_{j=1}^M$. To properly indicate the relative importance of each view/line, we additionally employ a 3D feature transformation branch independent of the $M$ 2D-guided local aggregation. This non-parametric branch only contains the local grouping in 3D space, feature average pooling within local groups, and propagation operations, converting the 3D token before the adapter into a feature baseline for adaptive ensemble, denoted as $B_i$. Then, we calculate the relative weights for different views/lines by the cosine similarity, and finally aggregate their features to obtain the final output as
\begin{equation}
     T_{i}^{out} = \frac{1}{M}\sum_{j=1}^M \operatorname{Sim}(B_i, F_{ij}).
\end{equation}
With the ensemble strategy, we integrate $M$ different features with dynamic weights, enabling the adapter to adaptively determine which view/line is more critical, contributing to high-quality adapted features.

\section{Experiments}
\label{sect:Experiments}

% In Section~\ref{sect:Settings}, we first introduce our fine-tuning settings on the ScanObjectNN~\cite{uy2019revisiting} and ModelNet40~\cite{wu20153d} datasets. Then in Section~\ref{sect:Quantitative}, we present the experiment results of Any-Modality Transformers (language, 2D image and audio) transferred to the 3D classification tasks. Finally in Section~\ref{sect:Ablation}, we conduct ablation studies to
% investigate the characteristics of the Any2Point framework.

In this section, we conduct extensive experiments on the ScanObjectNN~\cite{uy2019revisiting} and ModelNet40~\cite{wu20153d} datasets. We first introduce the fine-tuning settings and implementation details in Section~\ref{sect:Settings}. Then, in Section~\ref{sect:Quantitative}, we present the main experiment of transferring any-modality large models (language, 2D image and audio) to 3D classification tasks. Finally, in Section~\ref{sect:Ablation}, we conduct ablation studies to evaluate each component within our proposed Any2Point framework.

\subsection{Experimental Settings}
\label{sect:Settings}
\paragraph{\textbf{ScanObjectNN.}}

The ScanObjectNN dataset~\cite{uy2019revisiting} consists of real-world 3D object scans, categorized into 15 distinct classes. We select the most challenging PB-T50-RS split to test the performance of the Any2Point framework without the voting strategy. For all models, we employ the AdamW optimizer~\cite{loshchilov2017decoupled} and the CosineAnnealing scheduler. The initial learning rate is set to 5e-4, with a weight decay factor of 0.05. We fine-tune the model for 300 epochs with a batch size of 32. For data augmentation, we use Random scaling, translation, and rotation. For language, 2D vision, and audio modalities, we respectively select the CLIP Text Encoder~\cite{radford2021learning}, DINO V2~\cite{oquab2023dinov2}, and ImageBind Audio Encoder~\cite{girdhar2023imagebind} as pre-trained models. For these three models, the transformer architecture is the same: a 12-block encoder with 768 feature channels and 1,024 input point number. The hyperparameter M in the 3D-to-any Virtual Projection is set to 6 with identical angles for the Any-Modality Transformers.
To match the shape of the original PEs within pre-trained models, we virtually project 3D points into a 1D line segment of length 77 with a line size of 2 in the language modality; a 2D plane measuring 512x512 with a patch size of 26 in the 2D vision modality; and a 2D plane sized 192x304 with a patch size of 16 in the audio modality.

\paragraph{\textbf{ModelNet40.}}
The ModelNet40 dataset~\cite{wu20153d} consists of 40 categories of synthesized 3D CAD models, with 9,843 training samples and 2,468 test samples. In our experiments on ModelNet40, we adopt the same fine-tuning settings and the same pre-trained models as in ScanObjectNN. For data augmentation, we utilize default random scaling and translation. Notably, during the testing process, we do not employ the voting strategy.

\begin{table}[tb]
\vspace{-0.1cm}
  \caption{\textbf{Comparisons on accuracy with previous methods on 3D classification datasets.} We report the pre-training modality (Pre-train), the number of learnable parameters (\#Param) on the "PB-T50-RS" split of ScanObjectNN (SCAN.) and ModelNet40 (MN.).{$^\dagger$} indicates utilizing the voting strategy.}
  \label{tab:results}
  \centering
  \begin{tabular}{l|c|c|c|c}
    \toprule
        \makecell*[c]{\textbf{Method}} &\textbf{Pre-train} &\textbf{\#Param(M)} &\textbf{SCAN.(\%)} &\textbf{MN.(\%)}\\
        \cmidrule(lr){1-1} \cmidrule(lr){2-2} \cmidrule(lr){3-3} \cmidrule(lr){4-4} \cmidrule(lr){5-5}
        % \color{gray}{\textit{Training from Scratch}}\vspace{0.06cm}\\
        Point-NN~\cite{zhang2023starting} &N/A &0.0 &64.9 &81.8\\
        PointNet~\cite{qi2017pointnet} &N/A &3.5 &68.0 &89.2\\
        % SpiderCNN~\cite{xu2018spidercnn}  &77.1 &79.5 &73.7\\
        PointNet++~\cite{qi2017pointnet++} &N/A &1.5 &77.9 &90.7\\
        DGCNN\cite{wang2019dynamic} &N/A &1.8 &78.1 &92.9\\
        % PointCNN &0.6 &78.5\\
        % BGA-PN++ &1.5 &80.2\\
        % GBNet &8.8 &80.5\\
        % PRANet &2.3 &81.0\\
        PointMLP~\cite{ma2022rethinking} &N/A &12.6  &85.4 &94.1\\
        Point-PN~\cite{zhang2023starting} &N/A &0.8 &87.1 &93.8\\
        PointNeXt~\cite{qian2022pointnext} &N/A &1.4  &87.7 &94.0\\
        % BGA-DGCNN~\cite{} &- &- &79.7 \\
        % BGA-PN++~\cite{} &- &- &80.2 \\
        % DRNet~\cite{} &- &- &80.3\\
        % GBNet~\cite{} &- &-  &80.5\\
        % SimpleView~\cite{} &- &-  &80.5\\
        % MVTN~\cite{} &- &-  &82.8\\
        % PointMLP~\cite{} &- &-  &85.2\\
        \midrule
        % \color{gray}{\textit{Self-supervised Pre-training}}\vspace{0.06cm}\\
        Point-BERT~\cite{yu2022point} &3D &22.1 &83.1 &92.7\\
        \quad w/ Point-PEFT~\cite{tang2024point} &3D  &0.6 &85.0 &93.4\\
        Point-MAE~\cite{pang2022masked} &3D &22.1 &85.2 &93.2\\
        Point-M2AE~\cite{zhang2022point} &3D &15.3 &86.4 &93.4\\
        \midrule
        P2P-HorNet~\cite{wang2022p2p} &2D &1.2 &89.3 &94.0{$^\dagger$}\\
        ACT~\cite{dong2022autoencoders} &3D+2D &22.1 &88.2 &93.7\\
        I2P-MAE~\cite{zhang2023learning} &3D+2D &12.9 &90.1 &93.7\\
        ReCon~\cite{qi2023contrast} &3D+2D+Language &43.6 &90.6 &94.1\\
        \midrule
        \multirow{3}*{Any2Point} &Audio &\textbf{0.8} &\textbf{87.0} &\textbf{92.7}\\
        &2D &\textbf{0.8} &\textbf{87.7} &\textbf{93.2}\\
        &Language &\textbf{0.9} &\textbf{91.9} &\textbf{94.3}\\

  \bottomrule
  \end{tabular}
  \vspace{-0.4cm}
\end{table}

\subsection{Quantitative Analysis}
\label{sect:Quantitative}

The results are shown in Table~\ref{tab:results}. It is observed that: (i) On the 3D real-world object dataset ScanObjectNN, the Any2Point framework achieves 91.9\%, 87.7\%, and 87.0\% accuracy based on Language (CLIP-Text), 2D Vision (DINO V2-B), and Audio (ImageBind-Audio) modalities, respectively. Compared to the previous SOTA method (ReCon), 1D language pre-trained Any2Point achieves a 1.3\% improvement with only 0.9M learnable parameters. For the 2D (Vision/Audio) modalities, Any2Point significantly outperforms Point-M2AE, which is the SOTA method pre-trained only on 3D datasets, by 0.6\% and 1.3\%, respectively. This reveals that our framework is capable of fully exploiting pre-trained knowledge from other modalities to solve 3D recognition tasks.
(ii) On the 3D synthetic object dataset ModelNet40, across the Language, 2D Vision, and Audio modalities, our Any2Point framework attains 94.3\%, 93.2\%, and 92.7\%. 
Our framework exclusively utilizes one pre-trained model in the 1D language modality, achieving a 0.2\% improvement over the previous SOTA method (ReCon), and reducing 42.7M learnable parameters.
For 2D modalities, Any2Point demonstrates performance on par with models pre-trained exclusively on 3D datasets. 
(iii) Surprisingly, whether on the ScanObjectNN or the ModelNet40 dataset, the Any2Point framework maintains a performance trend where 1D modality (language) outperforms 2D modalities (image and audio). Large language models provide abundant spatial and semantic information in low-dimensional spaces to assist in 3D learning. This trend is further validated in the upcoming Section~\ref{sect:Ablation}.

\subsection{Ablation Study}
\label{sect:Ablation}
In this section, we conduct extensive ablation studies to explore the effectiveness of different components within our Any2Point framework. We adopt CLIP-Text (1D) and DINO V2 (2D) as the pre-trained transformer, and report the classification accuracy (\%) on the "PB-T50-RS" split of the ScanObjectNN dataset.

\begin{table}[t!]
  \caption{\textbf{Ablation Study on Different PEFT Methods.} We report the number of learnable parameters (\#P) and classification accuracy(\%) of CLIP-Text (1D.) and DINO V2 (2D.) on the "PB-T50-RS" split of the ScanObjectNN dataset.}
  \label{tab:ablation_PEFT}
  \centering
  \begin{tabular}{l|c|c|c}
    \toprule
        \makecell*[c]{\textbf{Method}} &\textbf{\#P(M)} &\textbf{1D.(\%)} &\textbf{2D.(\%)}\\
        \cmidrule(lr){1-1} \cmidrule(lr){2-2} \cmidrule(lr){3-3} \cmidrule(lr){4-4}
        % \color{gray}{\textit{Training from Scratch}}\vspace{0.06cm}\\
        Full Fine-Tuning &86.3 &79.9 &85.3\\
        \midrule
        Prompt Tuning~\cite{jia2022visual} &0.4 &89.1 &86.4\\
        Adapter Tuning~\cite{houlsby2019parameter} &0.4 &89.6 &85.9\\
        LoRA~\cite{hu2021lora} &0.9 &86.3 &85.1\\
        \midrule
        Any2Point &\textbf{0.8} &\textbf{91.9} &\textbf{87.7}\\

  \bottomrule
  \end{tabular}
\end{table}

\begin{table}[tb]
\vspace{-0.1cm}
  \caption{\textbf{Ablation Study on Main Components.} 
To validate the effectiveness of 3D-to-any Virtual Projection (V.P.) and Any-to-3D Guided Adapter (G.A.).}
  \label{tab:ablation_main}
  \centering
  \begin{tabular}{c|c|c|c|c}
    \toprule
        \makecell*[c]{3D-to-any V.P.} &\makecell*[c]{Any-to-3D G.A.} &\textbf{\#P(M)} &\textbf{1D.(\%)} &\textbf{2D.(\%)}\\
		 \cmidrule(lr){1-1} \cmidrule(lr){2-2} \cmidrule(lr){3-3} \cmidrule(lr){4-4} \cmidrule(lr){5-5}
        -&-&0.3&88.7&86.1\\
        % \rowcolor{gray!12}
        \checkmark&-&0.3&89.3&86.6\\
        % \cmidrule(lr){1-7}
        -&\checkmark&0.8&90.9&87.6\\
        \checkmark &\checkmark &\textbf{0.8} &\textbf{91.9} &\textbf{87.7}\\
  \bottomrule
  \end{tabular}
\vspace{-0.2cm}
  
\end{table}

\paragraph{\textbf{Comparison with traditional PEFT methods.}}
As demonstrated in Table \ref{tab:ablation_PEFT}, our Any-to-3D Guided Adapter significantly outperforms traditional PEFT techniques when utilizing pre-trained models from 1D or 2D modalities. 
In comparison to Prompt Tuning~\cite{jia2022visual}, it achieves improvements of 2.8\% and 1.3\%; compared to Adapter Tuning~\cite{houlsby2019parameter}, it achieves improvements of 2.3\% and 1.8\%; and in contrast to Low-Rank Adaptation (LoRA)~\cite{hu2021lora}, it achieves improvements of 5.6\% and 2.6\%, respectively.  
% Through extensive experiments, we have proven the effectiveness of the Any2Point framework over traditional PEFT methods in the 1D/2D domains for 3D perception tasks, efficiently mining and integrating pre-trained knowledge from other modalities into the understanding of 3D object semantics. Unlike other methods, our framework fully exploits the spatial knowledge from source modalities to capture the local fine-grained structural information of 3D objects.
The experimental results demonstrate that our Any-to-3D Guided Adapter can efficiently mine and integrate pre-trained knowledge from other modalities to understand the semantics of 3D objects. Unlike other methods, our framework leverages 1D/2D spatial guidance to aggregate the local semantics of 3D tokens, capturing the local fine-grained information of 3D objects.

\paragraph{\textbf{Effectiveness of Main Components.}}

As shown in Table~\ref{tab:ablation_main}, to substantiate the efficacy of our proposed methods, we conducted ablation experiments by progressively incorporating each component into the baseline. The first row indicates the baseline configuration, which consists of the 3D tokenizer, the pre-trained transformer, and the task head, with updates applied only to the tokenizer and head. Introducing the 3D-to-any Virtual Projection resulted in performance improvements to 89.3\% in the 1D modality and 86.6\% in the 2D modality. This suggests that using virtual projection, rather than true projection, helps mitigate the loss of 3D spatial information caused by modality conversion. Following the inclusion of the Any-to-3D Guided Adapter, performance in the 1D modality surged to 90.9\%, while in the 2D modality, it rose to 87.6\%, with a focus on local structures leading to greater improvements. Introducing both aforementioned methods simultaneously led to a surge in performance to 91.9\% in the 1D modality and a rise to 87.7\% in the 2D modality, effectively showcasing the effectiveness of our comprehensive framework.

\begin{table}[t!]
  \caption{\textbf{Ablation Study on 3D-to-any Virtual Projection.} Sinusoidal, Learnable and 3D-to-any V.P. refer to sinusoidal positional encoding, learnable positional encoding and 3D-to-any Virtual Projection.}
  \label{tab:VP}
  \centering
  \begin{tabular}{c|c|c|c|c}
    \toprule
        \makecell*[c]{Sinusoidal} &\makecell*[c]{Learnable} &\makecell*[c]{3D-to-any V.P.} &\textbf{1D.(\%)} &\textbf{2D.(\%)}\\
		 \cmidrule(lr){1-1} \cmidrule(lr){2-2} \cmidrule(lr){3-3} \cmidrule(lr){4-4} \cmidrule(lr){5-5}
        -&-&-&90.9&87.6\\
        % \rowcolor{gray!12}
        \checkmark&-&-&87.4&86.0\\
        % \cmidrule(lr){1-7}
        -&\checkmark&-&90.5&86.5\\
        -&-&\checkmark &\textbf{91.9} &\textbf{87.7}\\
  \bottomrule
  \end{tabular}
\end{table}

\begin{table}[tb]
  \caption{\textbf{Ablation Study on Any-to-3D Guided Adapter.}
 To validate the effectiveness of 1D/2D-guided Local Aggregation (L.A.) and Adaptive (Ada.) Any-to-3D Ensemble (Ens.).
 % 1D/2D-guided L.A. stands for 1D/2D-guided Local Aggregation, and Ada. Any-to-3D Ens. represents Adaptive Any-to-3D Ensemble.
 }
  \vspace{-0.1cm}
  \label{tab:Adapter}
  \centering
  \begin{tabular}{c|c|c|c|c}
    \toprule
        \makecell*[c]{1D/2D-guided L.A.} &\makecell*[c]{Ada. Any-to-3D Ens.} &\textbf{\#P(M)} &\textbf{1D.(\%)} &\textbf{2D.(\%)}\\
		 \cmidrule(lr){1-1} \cmidrule(lr){2-2} \cmidrule(lr){3-3} \cmidrule(lr){4-4} \cmidrule(lr){5-5}
        -&-&0.25&89.3&86.6\\
        % \rowcolor{gray!12}
        \checkmark&-&0.8&90.2&86.8\\
        \checkmark &\checkmark &\textbf{0.8} &\textbf{91.9} &\textbf{87.7}\\
  \bottomrule
  \end{tabular}
\end{table}

\begin{table}[t!]
  \caption{\textbf{More Results on ScanObjectNN.}}
  \vspace{-0.1cm}
  \label{tab:More}
  \centering
  \begin{tabular}{l|c|c|c|c}
    \toprule
        \makecell*[l]{\textbf{Method}} & \textbf{Pre-train} & \textbf{Model} &\textbf{\#Param(M)} &\textbf{SCAN.(\%)}\\
        \cmidrule(lr){1-1} \cmidrule(lr){2-2} \cmidrule(lr){3-3} \cmidrule(lr){4-4} \cmidrule(lr){5-5}
        \multirow{3}*{Any2Point} &Audio&SSAST~\cite{gong2022ssast}&\textbf{0.8} &\textbf{87.1}\\
        &2D& DeiT~\cite{touvron2021training} &\textbf{0.8} &\textbf{87.3}\\
        &Language& RoBERTa~\cite{liu2019roberta} &\textbf{0.9} &\textbf{89.7}\\

  \bottomrule
  \end{tabular}
\end{table}

\paragraph{\textbf{Effects of 3D-to-any Virtual Projection.}}
% In Table~\ref{tab:VP}, we investigate the effects of 3D-to-any Virtual Projection on the overall Any2Point framework. The first row indicates the absence of any positional encoding. Incorporating the sinusoidal positional encoding or learnable positional encoding leads to a certain degree of performance degradation. This is because the newly added positional information conflicts with the inherent semantics within the Any-Modality Transformer. On the other hand, employing 3D-to-any Virtual Projection yields a gain of 1.6\% and 0.1\% for 1D and 2D modalities, suggesting that the positional encoding obtained from virtual projection is well-aligned with the source modality pre-trained model.
In Table~\ref{tab:VP}, we investigated the effects of employing different positional encoding methods on the Any2Point framework. The first row indicates the absence of any positional encoding. Introducing sinusoidal positional encoding or learnable positional encoding led to a certain degree of performance degradation. This is due to the conflict between the newly introduced positional information and the inherent semantics within the source modality transformer. On the other hand, employing 3D-to-any Virtual Projection resulted in respective improvements of 1.0\% and 0.1\% accuracy. The results demonstrate that using original 1D/2D positional priors can promote the pre-trained transformer to acquire 3D features.

\paragraph{\textbf{Components of Any-to-3D Guided Adapter.}}
As shown in Table~\ref{tab:Adapter}, we conduct ablation experiments by incrementally adding components to the Any-to-3D Guided Adapter. The first row signifies the baseline adapter, consisting of only an MLP with bottleneck layers. By incorporating 1D/2D-guided Local Aggregation, composed of local aggregation in 1D/2D spaces, self-attention interactions, pooling, and propagation, our approach achieves performance gains of 0.9\% and 0.2\%. Leveraging the positional priors from the pre-trained model facilitates mining fine-grained 3D structural information from different perspectives. The Adaptive Any-to-3D Ensemble brings further improvements of 1.7\% and 0.9\% for 1D and 2D modalities, effectively integrating parallel features in accordance with 3D structural features. The experiments demonstrate the effectiveness of each component in our Any-to-3D Guided Adapter to gather 3D local geometric information, complementing the global attention in the pre-trained model.

\paragraph{\textbf{More Results on Performance Trend.}}
To further validate our previous findings that the Any2Point framework, based on 1D Language pre-trained models, significantly outperforms those based on 2D modalities (Vision/Audio) in the 3D object recognition task, we conduct additional experiments in Table~\ref{tab:More}. On the "PB-T50-RS" split of ScanObjectNN dataset, we select RoBERTa (1D), DeiT (2D Vision), and SSAST (Audio) as the pre-trained models, with fine-tuning settings consistent with our previous experiments. These models achieve performance of 89.7\%, 87.3\%, and 87.1\%, respectively. The performance trend across modalities is observed: 1D language > 2D Vision > 2D Audio. We suspect that due to the pre-training data, large language models possess stronger semantic information compared to other modalities, which is beneficial for the deep understanding of different 3D objects.

\section{Visualization}
\label{sec:visualization}

In this section, we opt to validate the efficacy of the proposed 3D-to-any Virtual Projection and the Any-to-3D Guided Adapter by visualizing on the ScanObjectNN test set, utilizing the CLIP-Text Encoder (1D)~\cite{radford2021learning} and DINO V2 (2D)~\cite{oquab2023dinov2}.

\begin{figure}[t!]
  \centering
  \includegraphics[height=4.7cm,width=12.5cm]{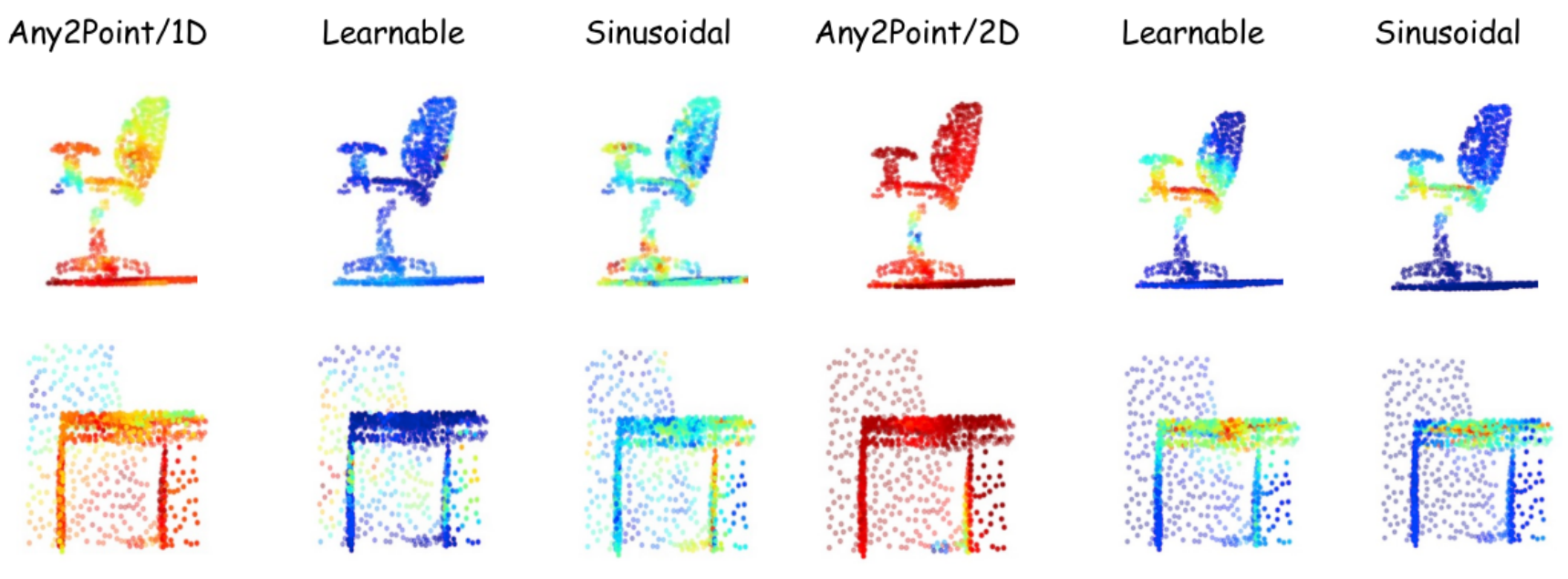}
  \caption{\textbf{Visualization of Different Positional Encoding Methods.} For the 1D/2D modalities, we visualize the attention scores of the \textbf{[CLS]} token to other point cloud tokens, utilizing sinusoidal positional encoding, learnable positional encoding, and 3D-to-any Virtual Projection. \textbf{The red color indicates higher values.}}
  \label{fig:vis1}
  \vspace{-0.4cm}
\end{figure}

\begin{figure}[tb]
  \centering
  \includegraphics[height=4.7cm,width=12.5cm]{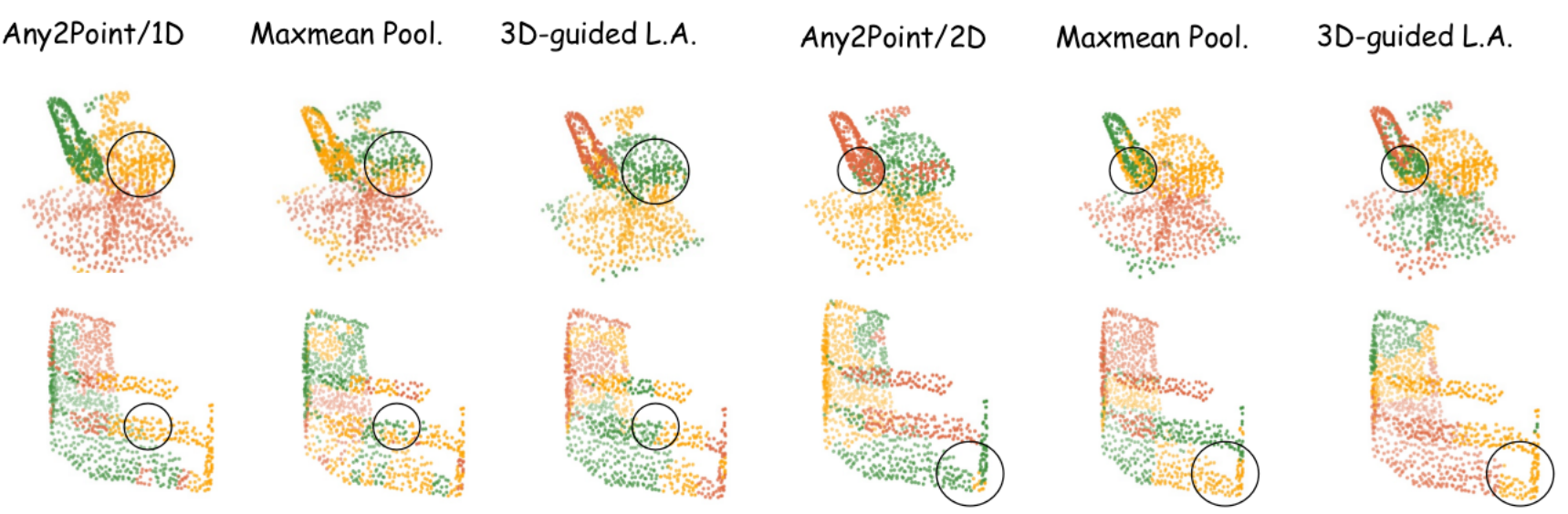}
  \caption{\textbf{Visualization of Effects of Any-to-3D Guided Adapter.}For 1D/2D modalities, we visualize the clusters of the similarities between the \textbf{[CLS]} token and other point features, with the number of clusters set to 3. It is conducted for the complete Any-to-3D Guided Adapter, for replacing the Adaptive Any-to-3D Ensemble with maxmean pooling (Maxmean Pool.), and for performing local aggregation solely based on 3D positions (3D-guided L.A.).}
  \label{fig:vis2}
  \vspace{-0.2cm}
\end{figure}

\subsection{Different Positional Encoding Methods}
Our 3D-to-any Virtual Projection fully exploits the positional encoding paired with the pre-trained model, injecting the source modality spatial knowledge into the 3D tokens during fine-tuning. In Figure~\ref{fig:vis1}, when using sinusoidal positional encodings, learnable positional encodings, and our 3D-to-any Virtual Projection respectively, we visualize the attention scores of the [CLS] token to other point cloud tokens. As illustrated, for the 1D language modality, learnable positional encodings grasp useless information. After applying the commonly used sinusoidal positional encodings in Large Language Models, they fail to capture the critical 3D semantics. However, our method focuses more on the salient object parts, such as the armrests and wheels of chairs, and the legs of tables. For the 2D visual modality, learnable encodings are slightly better than sinusoidal positional encodings, as 2D pre-trained models mainly adopt the learnable encoding method. Meanwhile, our method directly recognizes the whole object and its key parts, for example, giving high weights to the chair's backrest.

\subsection{Effects of Any-to-3D Guided Adapter}
The Any-to-3D Guided Adapter captures the 3D fine-grained information through interactions within the local regions of the source modality. In Figure~\ref{fig:vis2}, we visualize the clustering results of the similarities between the [CLS] token and other point token features, utilizing the complete Any-to-3D Guided Adapter, replacing the Adaptive Any-to-3D Ensemble with maxmean pooling, and further only using 3D positional information. As shown, for simple objects like chairs (1$^{st}$ row), our method effectively distinguishes between the chair's backrest, armrests, seat, and wheels, whereas removing components fails to capture the differences between key parts. For more challenging objects like shelves (2$^{nd}$ row), removing any components leads to semantic confusion of the object, while our approach clearly differentiates the shelf's base, middle layer, and backrest. These experiments indicate that each component within the Adapter effectively utilizes the positional information from different modalities to promote the 3D structure extraction.

% \begin{figure}[tb]
%   \centering
%   \begin{subfigure}{0.68\linewidth}
%     \fbox{\rule{0pt}{0.5in} \rule{.9\linewidth}{0pt}}
%     \caption{An example of a subfigure}
%     \label{fig:short-a}
%   \end{subfigure}
%   \hfill
%   \begin{subfigure}{0.28\linewidth}
%     \fbox{\rule{0pt}{0.5in} \rule{.9\linewidth}{0pt}}
%     \caption{Another example of a subfigure}
%     \label{fig:short-b}
%   \end{subfigure}
%   \caption{Centered, short example caption}
%   \label{fig:short}
% \end{figure}

\section{Additional Ablation Study}
\label{abla}
In this section, we conduct ablation studies to explore the effectiveness of different components and hyper-parameters, which are not discussed in the main text. We select CLIP-Text~\cite{radford2021learning} (1D) and DINO V2~\cite{oquab2023dinov2} (2D) as pre-trained models to compare the accuracy (\%) on the "PB-T50-RS" split of ScanObjectNN~\cite{uy2019revisiting}.

\begin{table}[h!]
\vspace{-0.1cm}
  \caption{\textbf{Ablation Study on Adapter Positions and Depths.} The `After', `Before', and `Parallel' denote block structure of putting the \textit{Any-to-3D Guided Adapter} after, before, and in parallel to the FFN layer respectively. The insertion depth denotes the number of earlier blocks. We report the classification accuracy of CLIP-Text~\cite{radford2021learning} (1D.) and DINO V2~\cite{oquab2023dinov2} (2D.) on the "PB-T50-RS" split of ScanObjectNN~\cite{uy2019revisiting}.}
  \label{tab:depth}
  \centering
  \begin{tabular}{cccccc}
    \toprule
    \multicolumn{3}{c}{\ \ \ \ \ \ \ \ \ Positions of Any-to-3D Guided Adapter\ \ \ \ \ \ \ \ \ } &\makecell*[c]{\multirow{2}*{\shortstack{\vspace*{2.2pt}\\Insertion\\\vspace*{0.3pt}\\Depth}}} &\makecell*[c]{\multirow{2}*{\shortstack{\vspace*{2.2pt}\\1D.\\\vspace*{0.3pt}\\(\%)}}} &\makecell*[c]{\multirow{2}*{\shortstack{\vspace*{2.2pt}\\2D.\\\vspace*{0.3pt}\\(\%)}}} \\
    \cmidrule(lr){1-3}
      \ \ \ \ After\ \ \ \  &\ \ Before\ \ \ &\ \ Parallel\ \ \  & &\\
     \cmidrule(lr){1-1}  \cmidrule(lr){2-2}  \cmidrule(lr){3-3}  \cmidrule(lr){4-4}  \cmidrule(lr){5-5} \cmidrule(lr){6-6} 
     \checkmark &- &-  &3 &90.4 &86.9 \\
     \checkmark &- &-  &6 &90.6 &87.1 \\
     \checkmark &- &-  &9 &90.2 &87.2 \\
     \checkmark &- &-  &12 &\textbf{91.9} &\textbf{87.7} \\
     \cmidrule(lr){1-3}
     - &\checkmark &-  &3 &89.9 &86.2\\
     - &\checkmark &-  &6 &91.1 &86.4\\
     - &\checkmark &-  &9 &90.4 &86.4\\
     - &\checkmark &-  &12 &91.3 &87.0\\
     \cmidrule(lr){1-3}
     - &- &\checkmark  &3 &90.3 &86.7\\
     - &- &\checkmark  &6 &91.5 &86.8\\
     - &- &\checkmark  &9 &90.4 &86.8\\
     - &- &\checkmark  &12 &90.7 &87.1\\
    \bottomrule
  \end{tabular}
  \vspace{-0.4cm}
\end{table}

\subsection{The Adapter Positions and Depths.}

In Table~\ref{tab:depth}, we further conducted ablation studies on the positions and depths of the proposed Any-to-3D Guided Adapter. As shown in Table~\ref{tab:depth}, the best performance is achieved when the adapter is placed after the Feed Forward Networks (FFNs) and at a depth of 12. This is because, when placed after the FFN layers, the globally interacted point cloud features undergo local aggregation within the adapter, extracting the fine-grained structures of the 3D point clouds. Moreover, deeper insertion allows the adapter to leverage both low-level and high-level pre-trained knowledge to process the point cloud information. It is important to note that for all pre-trained models, we inserted the Any-to-3D Guided Adapter after the FFN layers of all blocks.

\begin{table}[t!]
\vspace{-0.1cm}
  \caption{\textbf{Ablation Study on the Components of 3D-to-any Virtual Projection.} The 2D Proj.V1 and 2D Proj.V2 denote the projection methods used in PointCLIP V1~\cite{zhang2022pointclip} and PointCLIP V2~\cite{zhu2023pointclip}, respectively. The 1D Proj. refers to the projection method depicted in the main text. Meanwhile, we ablate the projection view number (Num.) on the pre-trained models.}
  \label{tab:view}
  \centering
  \begin{tabular}{c|c|c|c|cc}
    \toprule
     \makecell*[c]{2D Proj.V1} &\makecell*[c]{2D Proj.V2}  &\makecell*[c]{1D Proj.} &\makecell*[c]{ View Num.} &\makecell*[c]{1D. (\%)} &\makecell*[c]{2D. (\%)} \\
     \cmidrule(lr){1-1}  \cmidrule(lr){2-2}  \cmidrule(lr){3-3}  \cmidrule(lr){4-4}  \cmidrule(lr){5-5} \cmidrule(lr){6-6} 
     \checkmark &- &-  &4 &- &86.2 \\
     \checkmark &- &-  &6 &- &\textbf{87.7} \\
     \checkmark &- &-  &8 &- &87.2 \\
     \cmidrule(lr){1-4}
     - &\checkmark &-  &4 &- &86.4\\
     - &\checkmark &-  &6 &- &87.1\\
     - &\checkmark &-  &8 &- &86.5\\
     \cmidrule(lr){1-4}
     - &- &\checkmark  &4 &90.6 &-\\
     - &- &\checkmark  &6 &\textbf{91.9} &-\\
     - &- &\checkmark  &8 &89.8 &-\\
    \bottomrule
  \end{tabular}
  \vspace{-0.4cm}
\end{table}

\subsection{The Components of 3D-to-any Virtual Projection.}

In Table~\ref{tab:view}, we validated the impact of different projection methods and the number of projection views on the pre-trained models for different modalities. As shown in Table~\ref{tab:view}, for the 1D/2D modalities, the optimal performance is obtained when the number of views is set to 6. Meanwhile, for the 2D modality, the simple projection in PointCLIP~\cite{zhang2022pointclip} performs better than the more complex projection in PointCLIP V2~\cite{zhu2023pointclip}. The findings suggest that employing an appropriate number of projection views sufficiently captures the diversity and complexity inherent in 3D data in low-dimensional spaces. Furthermore, a simple projection method proves adequate for representing the fine-grained structure and data characteristics of 3D point clouds. It is worth noting that we use simple projection~\cite{zhang2022pointclip} and 6 views for all pre-trained models of any modalities.

\begin{table}[tb]
\vspace{-0.1cm}
  \caption{\textbf{Ablation Study on the Different Local Neighborhood Sizes.} We conduct ablation experiments for different combinations of 1D line sizes, 2D patch sizes, and 3D grid sizes.}
  \label{tab:size}
  \centering
  \begin{tabular}{c|c|c|cc}
    \toprule
     \makecell*[c]{Patch Size} &\makecell*[c]{Line Size} &\makecell*[c]{ Grid Size} &\makecell*[c]{1D. (\%)} &\makecell*[c]{2D. (\%)} \\
     \cmidrule(lr){1-1}  \cmidrule(lr){2-2}  \cmidrule(lr){3-3}  \cmidrule(lr){4-4}  \cmidrule(lr){5-5} 
     16 &-  &0.08 &- &86.5 \\
     26 &-  &0.08 &- &87.2 \\
     34 &-  &0.08 &- &86.4 \\
     \cmidrule(lr){1-5}
     16 &-  &0.16 &- &87.0\\
     26 &-  &0.16 &- &\textbf{87.7}\\
     34 &-  &0.16 &- &86.0\\
     \cmidrule(lr){1-5}
     - &1   &0.08 &90.9 &-\\
     - &2   &0.08 &\textbf{91.9} &-\\
     - &3   &0.08 &90.8 &-\\
     \cmidrule(lr){1-5}
     - &1   &0.16 &88.9 &-\\
     - &2   &0.16 &89.3 &-\\
     - &3   &0.16 &91.1 &-\\
    \bottomrule
  \end{tabular}
  \vspace{-0.1cm}
\end{table}

\begin{table}[t!]
\vspace{0.1cm}
  \caption{\textbf{Ablation Study on the 1D/2D-guided Local Aggregation.} To further emphasize the significance of 1D/2D-guided Local Aggregation (L.A.), we conduct additional experiments on the "PB-T50-RS" split of the ScanObjectNN~\cite{uy2019revisiting} dataset.}
  \label{tab:agg}
  \centering
  \begin{tabular}{c|c|c|cc}
    \toprule
     \makecell*[c]{3D-guided L.A.} &\makecell*[c]{2D-guided L.A.} &\makecell*[c]{1D-guided L.A.} &\makecell*[c]{1D. (\%)} &\makecell*[c]{2D. (\%)} \\
     \cmidrule(lr){1-1}  \cmidrule(lr){2-2}  \cmidrule(lr){3-3}  \cmidrule(lr){4-4} \cmidrule(lr){5-5}  
     \checkmark &- &- &- &86.2 \\
     - &\checkmark &- &- &\textbf{87.7} \\
     \cmidrule(lr){1-5}  
     \checkmark &- &-  &87.4 &- \\
     - &- &\checkmark &\textbf{91.9} &- \\
    \bottomrule
  \end{tabular}
  \vspace{-0.4cm}
\end{table}

\subsection{The Influences of Different Local Neighborhood Sizes.}

In Table~\ref{tab:size}, we investigated the performance impact of various combinations of 1D line sizes, 2D patch sizes, and 3D grid sizes~\cite{wu2022point} during the 1D/2D-guided local aggregation stage. As demonstrated in Table~\ref{tab:size}, when the 1D line size and 2D patch size are set to moderate values of 2 and 26, respectively, remarkable performance is attained. These findings indicate that an appropriate local aggregation size enhances the model's comprehension of 3D local information, whereas excessively large or small sizes could lead to the loss of critical features. For diverse tasks, we have consistently employed similar local aggregation sizes for pre-trained transformers of 1D/2D modalities.

\begin{figure}[tb]
  \centering
  \includegraphics[height=8cm,width=12.5cm]{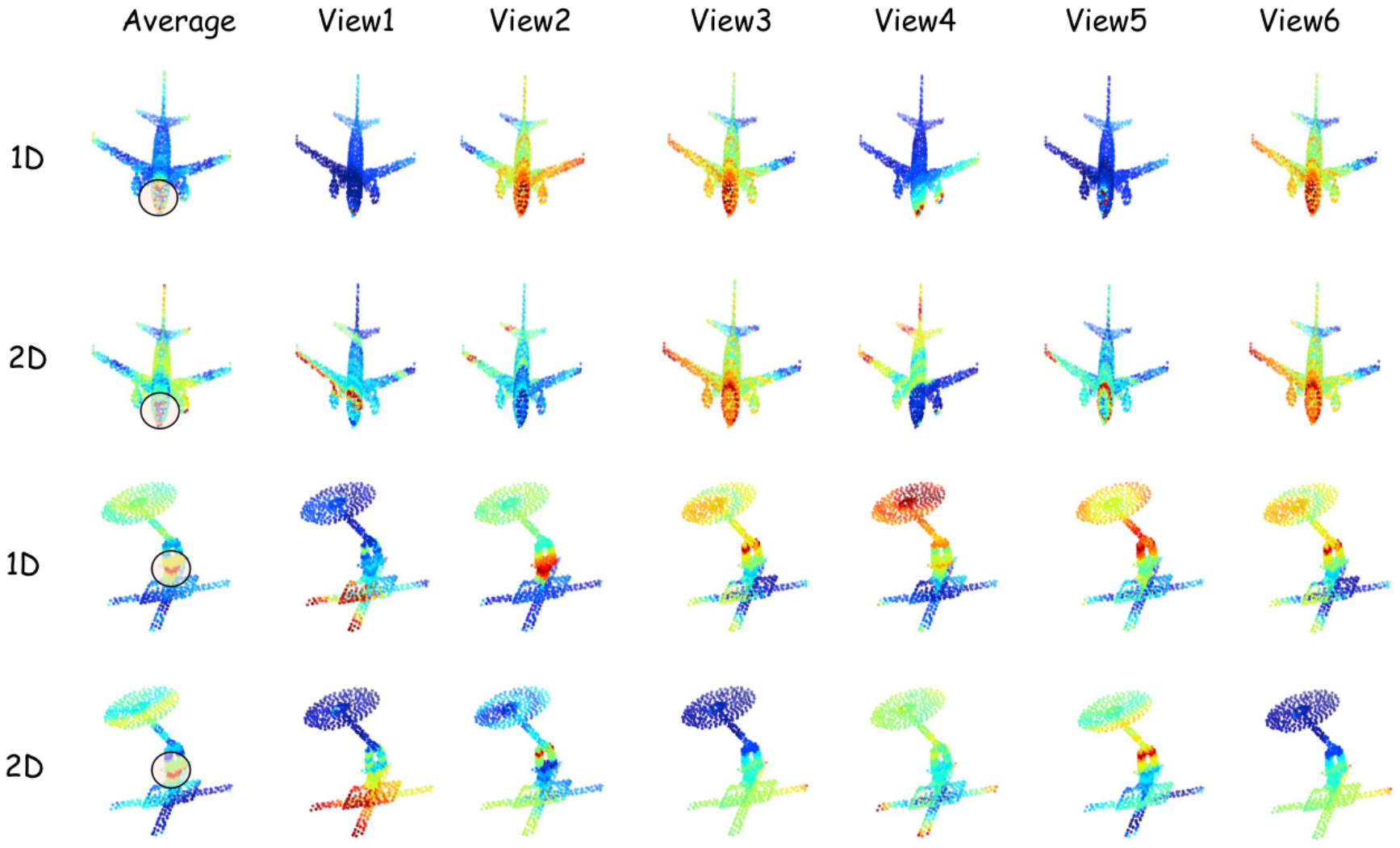}
  \caption{\textbf{Visualization of Different 3D Positional Embeddings.} For the 1D/2D modalities, we visualize the cosine similarity between the 3D positional embeddings (PEs) of selected point and those of other tokens on the airplane and lamp, respectively. The first column represents the averaging of 3D PEs obtained from 6 views, while the subsequent columns show the 3D PEs under different views. The black circle area in the first column represents the range where I select random point. \textbf{The red color indicates higher similarity.}}
  \label{fig:vis3}
  \vspace{-0.4cm}
\end{figure}

\subsection{The Importance of 1D/2D-guided Local Aggregation.}

In Table~\ref{tab:agg}, we validated the advantages of utilizing positional priors from the source modalities that are compatible with the pre-trained model, by injecting the positional priors of different modalities (1D/2D/3D) into the local aggregation step within the Any-to-3D Guided Adapter. As shown in Table~\ref{tab:agg}, compared to using the original 3D spatial knowledge, our method exhibits a performance improvement of 1.5\%-4.5\% for 2D/1D modalities. This demonstrates that our proposed method effectively addresses the issue of positional discrepancy between the original 3D positions and the pre-trained model.

\section{Additional Visualization}
\label{vis_}
We conduct visualization experiments on the test split of the ModelNet40 dataset~\cite{wu20153d} utilizing the CLIP-Text~\cite{radford2021learning} (1D) and DINO V2~\cite{oquab2023dinov2} (2D).

\subsection{The Significance of Encoding 3D Positions in 1D/2D PEs}

To demonstrate the effect of our method that uses 1D/2D Positional Embeddings (PEs) from source modalities to encode 3D positions, we randomly selected one point on a 3D object (inside the black circle in the first column) and computed the cosine similarity between the positional embedding of the point and those of other tokens. We compared 3D PEs assigned under different views (the six columns on the right) and the final 3D PEs obtained by averaging over M views (the first column), where M is set to 6. As shown in Figure~\ref{fig:vis3}, the first column shows higher similarity in areas closer to our selected point, while in other columns, high similarity areas are scattered at farther locations. For example, in the first and second rows, we selected a point on the nose of a plane, and in the third and fourth rows, we selected a point on the base of a lamp. In our method (the first column), similarity values decreases with distance, showing a transition from strong to weak, whereas in the other columns, the distribution appears irregular. It indicates that our proposed 3D positional embeddings implicitly establish spatial relationships in 3D space.

\section{Conclusion}
In conclusion, our paper proposes Any2Point to enable a general any-to-3D transferring framework, empowering any-modality pre-trained large models (e.g., 2D vision, language, and audio) for efficient 3D understanding. Within Any2Point framework, we introduce two techniques, named 3D-to-any virtual projection and any-to-3D guided adapter, to extract 3D structure knowledge while efficiently fine-tuning pre-trained models. This enables us to overcome issues within current methods, such as 3D geometry loss and excessive resource cost. Our extensive experiments across various tasks demonstrate the superior performance and efficiency of Any2Point compared to previous SOTA 3D pre-trained models, achieving remarkable results with only a fraction of the trainable parameters.

%\bibliographystyle{splncs04}
{
\bibliographystyle{plainnat}
\bibliography{main}
}
\end{document}